\documentclass{article}
\usepackage{ijcai23}

\usepackage{times}
\usepackage{soul}
\usepackage{url}
\usepackage[hidelinks]{hyperref}
\usepackage[utf8]{inputenc}
\usepackage[small]{caption}
\usepackage{graphicx}
\usepackage{amsmath}
\usepackage{amsthm}
\usepackage{booktabs}
\usepackage{algorithm}
\usepackage{algorithmic}
\usepackage[switch]{lineno}
\usepackage{xcolor} 
\usepackage{multirow}
\usepackage{amssymb}
\usepackage{pifont}
\title{StackFLOW: Monocular Human-Object Reconstruction by Stacked Normalizing Flow with Offset}

\author{
	Chaofan Huo$^{1,2}$
	\and
	Ye Shi$^{1,2}$\and
	Yuexin Ma$^{1,2}$\and
	Lan Xu$^{1,2}$\and
	Jingyi Yu$^{1,2}$\And
	Jingya Wang\thanks{Corresponding author.} $^{1,2}$\\
	\affiliations
	$^1$ShanghaiTech University\\
	$^2$Shanghai Engineering Research Center of Intelligent Vision and Imaging 
	\emails
	\{huochf, shiye, mayuexin, xulan1, yujingyi, wangjingya\}@shanghaitech.edu.cn,
}

\begin{document}
	
	\maketitle
	
	\begin{abstract}
	Modeling and capturing the 3D spatial arrangement of the human and the object is the key to perceiving 3D human-object interaction from monocular images. In this work, we propose to use the Human-Object Offset between anchors which are densely sampled from the surface of human mesh and object mesh to represent human-object spatial relation. Compared with previous works which use contact map or implicit distance filed to encode 3D human-object spatial relations, our method is a simple and efficient way to encode the highly detailed spatial correlation between the human and object. Based on this representation, we propose Stacked Normalizing Flow (StackFLOW) to infer the posterior distribution of human-object spatial relations from the image. During the optimization stage, we finetune the human body pose and object 6D pose by maximizing the likelihood of samples based on this posterior distribution and minimizing the 2D-3D corresponding reprojection loss. Extensive experimental results show that our method achieves impressive results on two challenging benchmarks, BEHAVE and InterCap datasets. Our code has been publicly available at \url{https://github.com/huochf/StackFLOW}.
\end{abstract}
	
	\section{Introduction}
	\begin{figure}[htbp]
	\centering
	\includegraphics[width=\linewidth]{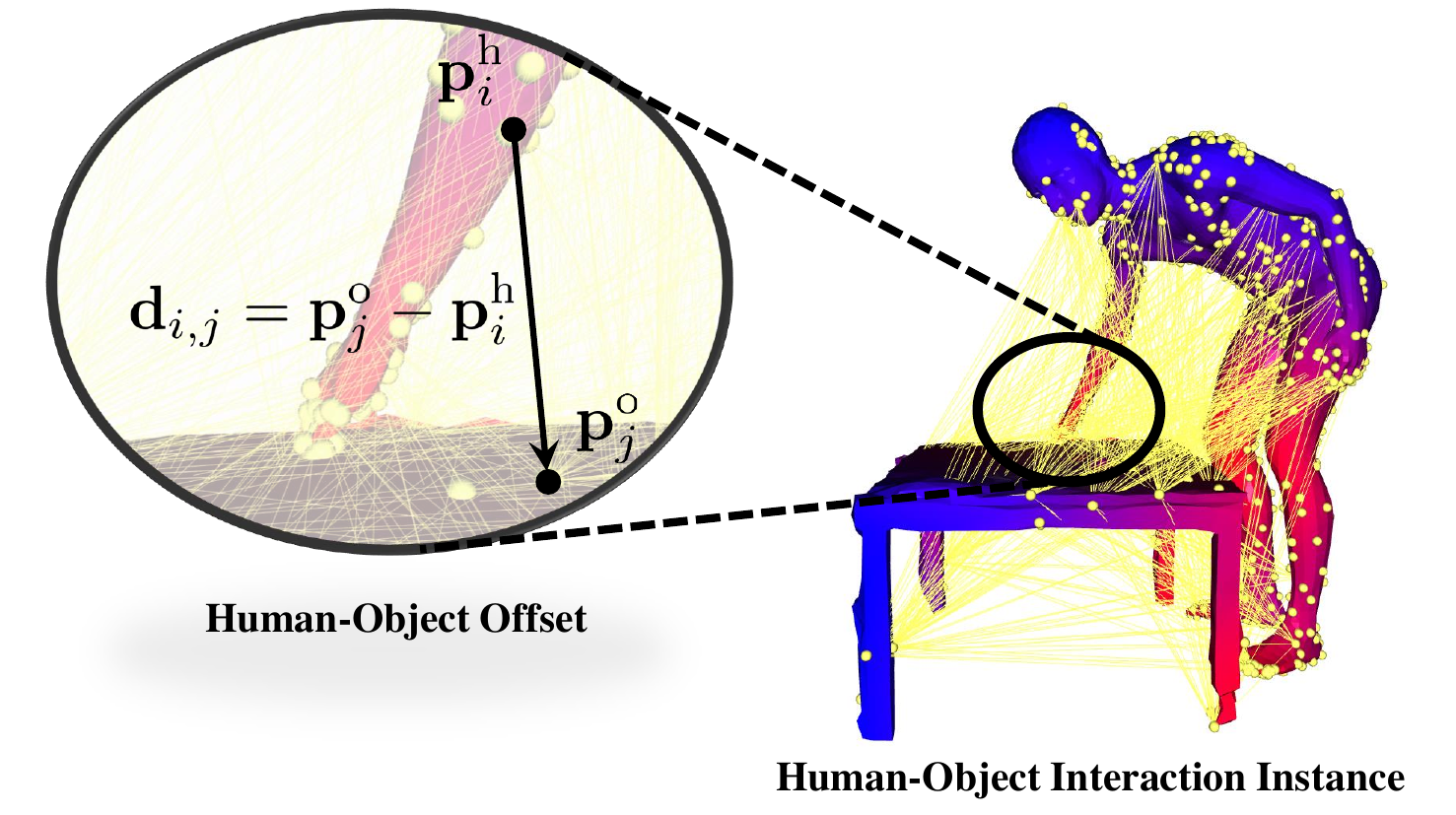}
	\caption{Human-Object Offset $\mathbf{d}_{i,j}$ describes how far from the human anchor point $\mathbf{p}_i^\text{h}$ to object anchor point $\mathbf{p}_j^\text{o}$ through the direction of the vector $\mathbf{d}_{i,j}$. They are calculated between two sets of anchors which are densely sampled from the surface of human mesh and object mesh beforehand. The dense offset captures a highly-detailed correlation between human parts and object parts. It is a quantitative representation to encode the 3D spatial relationship between the human and the object given human-object interaction instance. }
	\label{fig:ho_offset}
\end{figure}
	For a decade, the 3D information recovery of the human and the object from the image was studied alone, without considering their interaction. Recent studies suggest that the integration of humans and surrounding objects can produce physically consistent results \cite{Hassan_2019_ICCV:PROX}, and that the reconstruction accuracy of both can be improved even more \cite{Chen_2019_ICCV:holistic_plus,sun2021neural,Zhang_2023_CVPR:NeuralDome}.
In the monocular human-object reconstruction task, which aims at reconstructing human mesh and object mesh jointly from a single-view image, the interaction plays an important role in providing constraints for human pose and object position. However, how to utilize the interaction between the human and the object to refine themselves still remains unsolved.

The visible presentation of interaction between the human and the object in the 3D world is their spatial arrangement, which involves the posture of the human and the 6D pose of the object. Creating an appropriate representation for human-object spatial arrangement is vital for both human-object interaction capture from images and post-optimization for refinement. Contact map is a recently popular fine-grained representation to model the interaction between the human and the object. It has been applied to model human-scene interaction \cite{Huang_2022_CVPR:BSTRO} and human-object interaction \cite{zhang2020phosa,xie2022chore}. The contact map defines the contact regions in human mesh and object mesh and is suitable to be applied in the post-optimization step to generate plausible results by drawing closer contact points in human and object mesh. However, it only preserves the local contact information and cannot model the non-contact interaction types. Moreover, it relies on plausible initialization of the human and the object during optimization, and therefore it is not an independent representation to encode human-object spatial arrangement. Another way to encode human-object spatial arrangement is using the implicit distance field, which is a neural function that maps 3D points to point-to-face distances \cite{grasping_field,xie2022chore}. It is suitable to model 3D object shapes but some shortcomings may show up in modeling human-object spatial relationships. First, it is low-efficient since we need to sample many points to approximate the surface of the mesh. Moreover, spatial arrangement is encoded using functionalized representation implicitly rather than vectorized representation which results in applying probabilistic models to model the distribution of human-object spatial arrangement is difficult and indirect. In this paper, we pursue an efficient and unifying representation to encode the 3D spatial relationship between the human and the object. 

The relative distance frequently assumes a prominent role in numerous descriptions of 3D spatial relations, serving as a conspicuous and widely employed method of encoding. But in the scenario of human-object interaction, things may become complicated because of the articulated human body. In this work, we present a novel representation to encode human-object spatial relations using human-object offset. In order to involve all human body parts and cover various object shapes, we randomly sample anchor points from the surface of human mesh and object mesh. The offsets are calculated between all human anchors and object anchors for a given human-object pair as depicted in Figure \ref{fig:ho_offset}. We treat the offset as the numerical description of the spatial relation pattern for a target human-object pair. These offsets are representative since they encode highly detailed correlations between human parts and object parts. We can use them to recover the posture of the human and the position of the object by adjusting the position of human anchors and object anchors. Due to the regular topological structure of human mesh and rigid object mesh, these offsets are very redundant. Then we use PCA to transform these offsets from high-dimension offset space to low-dimension latent space by linear projection. The human-object offset is a generalized representation from the contact map since the contact map only keeps the anchors with zero offsets. 

Regressing accurate offsets from the image is hard due to the variety of spatial relations, the ambiguity of monocular capture, the indeterminacy of viewports, and the diversity of object scale. To tackle these problems, we design our method from two aspects. First, we use a probabilistic model to infer the distribution of spatial relationship instead of single point regression following \cite{Kolotouros_2021_ICCV:prohmr}. This distribution narrows down the search space of human-object spatial relations during the post-optimization step and more convincing results can be produced. Moreover, we decouple the process of inferring human-object spatial relation into two stacked subprocesses: human pose estimation and pose-conditioned distribution inference. With the guide of human pose, distribution for human-object spatial relations can be learned more stably and efficiently. 

Our contributions can be summarized as:
\begin{enumerate}
	\item  A new 3D spatial relation encoding technique is proposed to encode highly informative global correlation between the human and the object. The proposed Human-Object Offset (HO-offset) is densely sampled from the surface of human mesh and object mesh to construct latent spatial relation space.
	\item We propose a novel Stacked Normalizing Flow to infer the posterior distribution of human-object spatial relation for an input image. During inference, a new post-optimization process is designed with relative offset loss to constrain the body pose of the human and the 6D pose of the object.
	\item Our method outperforms the previous SOTA method with 16\% relative accuracy improvement and 88\% relative optimization time reduction.
\end{enumerate}

	\section{Related Works}
	\paragraph{Monocular 3D human-object reconstruction.} Although there are extensive works in 3D human mesh recovery \cite{Kanazawa_2018_CVPR,Kolotouros_2019_ICCV,Lin_2021_CVPR,liang2023hybridcap,Zhang_2023_AAAI} and 6D object pose estimation \cite{Kehl_2017_ICCV,Li_2019_ICCV,Chen_2022_CVPR}, reconstructing human and object jointly is yet a newly proposed problem. 3D human-object reconstruction can be divided into various settings, we only focus on reconstructing 3D human-object from a single-view RGB image. Towards reconstructing and understanding human activity in 3D scenes, \cite{Chen_2019_ICCV:holistic_plus} presents the 3D holistic scene understanding task, which combines 3D scene reconstruction and 3D human pose estimation. Physical commonsense about human-object interaction is utilized to improve the performance of these two tasks. Its follow-up work \cite{Weng_2021_CVPR} extends this direction to holistic human-object mesh reconstruction. They present an end-to-end trainable model that reconstructs both the human body and object meshes from a single RGB image. In the other direction, the scale of observed objects is zoomed in from the global human-scene to local human-object pairs. \cite{zhang2020phosa} tackles the problem of reconstructing human-object spatial arrangement in the wild. They propose an optimization-based framework which incorporates predefined 3D commonsense constraints to reduce the likely 3D spatial layout between the human and the object. \cite{xie2022chore} presents a unified data-fitting model that learns human-object spatial configuration priors from dataset \cite{Bhatnagar_2022_CVPR:BEHAVE} which is collected using multi-view capture systems. More recently, \cite{xie2023vistracker} tackles the challenge of single-view human-object tracking under heavy occlusion. Our work focuses on the second direction.

\paragraph{Spatial relationship modeling.} Modeling and capturing human-object spatial relationships are inescapable topics throughout various human-object interaction tasks. In the 2D human-object interaction detection task, spatial relation is encoded using relative 2D coordinates between the human bounding box and the object bounding box \cite{Gkioxari_2018_CVPR}, 2-channels mask map of human and object \cite{Ulutan_2020_CVPR}, relative locations between human parts and the center point of the object \cite{Wan_2019_ICCV} and two-direction spatial distributions between human body parts and object parts \cite{LIU2022108438}. Similarly, in 3D human-object reconstruction tasks it can be encoded using 3D space positions of human center and object center \cite{Chen_2019_ICCV:holistic_plus} or 3D relative spatial positions and orientations between object and person parts \cite{savva2016pigraphs}. But recently, more works admit that the contact map is a more fine-grained way to describe how humans and objects interact. \cite{zhang2020phosa} uses commonsense knowledge to define which parts in human and object mesh are participated in the interaction. \cite{xie2022chore} utilizes contact loss between human and object to get a more physically plausible and accurate reconstruction. This idea is also applied in human-scene interaction \cite{Huang_2022_CVPR:BSTRO}. Another popular way to model spatial relationships is using an implicit relative distance field. \cite{grasping_field} proposes the grasping field that is a continuous function mapping any points in 3D space to two point-to-surface signed distances. Towards hand-object reconstruction, they use a variational encoder-decoder network to learn it from data. This similar idea is also applied to human-scene interaction \cite{place} and human-object interaction \cite{xie2022chore}. Different from these works, we encode spatial relations using offset vectors between anchors densely sampled from the surface of human mesh and object mesh, while previous works just use the coarse relative distance between human parts and the center of the object or just focus on the local regions in contact. 

\paragraph{Probabilistic models in 3D reconstruction.} Due to the inherent ambiguity of monocular 3D reconstruction, probabilistic models are more appropriate for inferring distribution from partial observation rather than deterministic prediction. \cite{bui2020eccv} devises a multi-hypotheses method that continuously models the orientation of camera pose using Bingham distribution and camera position using multivariate Gaussian. \cite{Sengupta_2021_CVPR} infers multivariate Gaussian distribution of occluded or invisible body from a single image. Except for multivariate Gaussian, normalizing flow is another popular probabilistic model which is proposed in the context of variational inference \cite{pmlr-v37-rezende15} and density estimation \cite{real_nvp}. In the context of 3D reconstruction, more recent works utilize normalizing flow for human pose estimation \cite{Wandt_2022_CVPR}, human mesh recovery \cite{Kolotouros_2021_ICCV:prohmr}, two-hand reconstruction \cite{HandFlow_VMV2022}, conditioned human pose generation \cite{Aliakbarian_2022_CVPR} and human motion synthesis \cite{henter2020moglow}. Following these previous works, we deploy normalizing flow to learn the distribution of potential spatial arrangement between the human and the object from monocular images.

	\begin{figure*}[htbp]
	\centering
	\includegraphics[width=\linewidth]{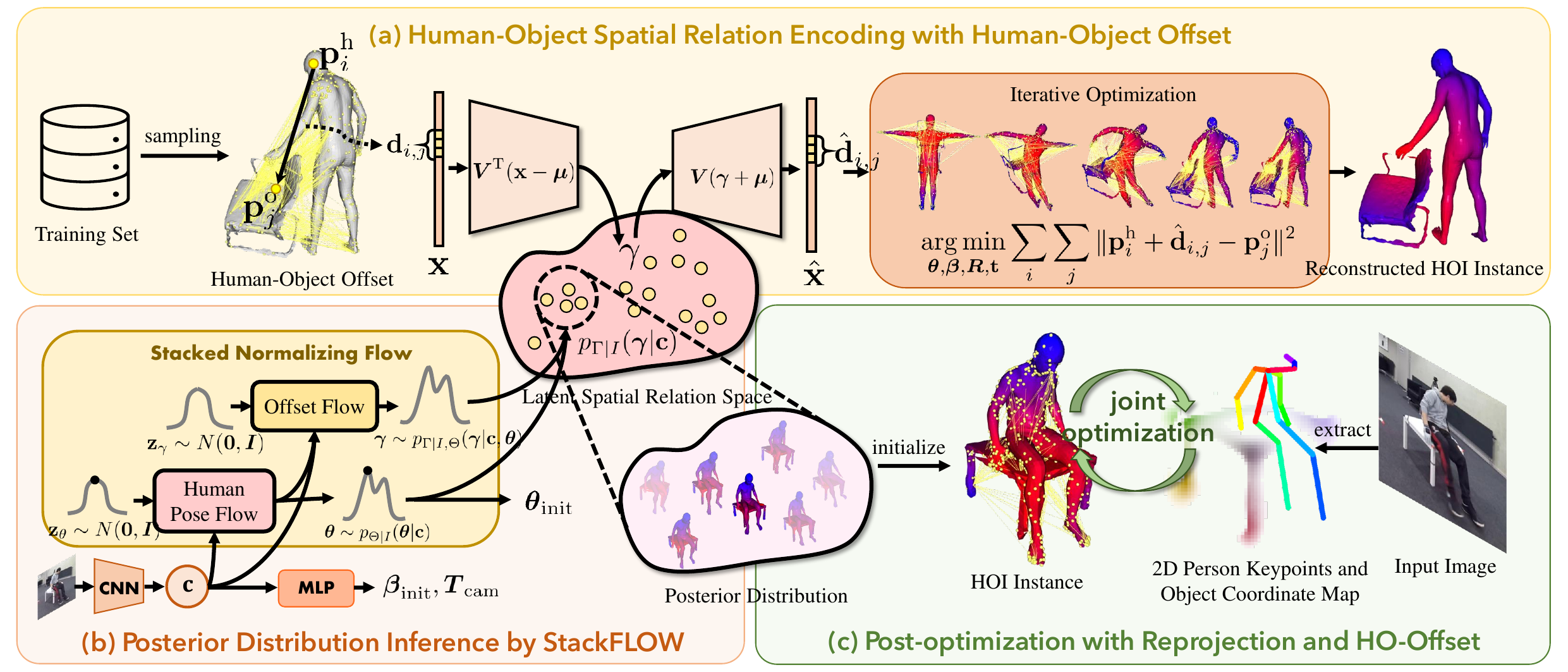}
	\caption{Main framework for our method. (a) We use human-object offset to encode the spatial relation between the human and the object. For a human-object pair, offsets are calculated and flattened into an offset vector $\mathbf{x}$. Based on all offset vectors calculated from training set, the latent spatial relation space is constructed using principle component analysis. To get a vectorized representation for human-object spatial relation, the offset vector is projected into this latent spatial relation space by linear projection. Inversely, given a sample $\gamma$ from this latent spatial relation space, we can reproject it to recover offset vector $\hat{\mathbf{x}}$. The human-object instance can be reconstructed from $\hat{\mathbf{x}}$ by iterative optimization. (b) With pre-constructed latent spatial relation space, we use stacked normalizing flow to infer the posteriori distribution of human-object spatial relation for an input image. The details are shown in Sec. \ref{section:distribution}. (c) In post-optimization stage, we further finetune the reconstruction results using 2D-3D reprojection loss and offset loss which is illustrated in Sec. \ref{section:optimization}.}
	\label{fig:main_pipeline}
\end{figure*}
	
	\section{Method}
	Given an input image and target object category, we aim at predicting the SMPL parameters including person shape $\boldsymbol{\beta}$, person pose $\boldsymbol{\theta}$ and object 6D pose i.e. rotation matrix $\boldsymbol{R}$ and translation $\mathbf{t}$. Since predicting these parameters in isolation will produce inconsistent results such as a floating object in the air or interpolation between the human and the object, we propose to use directed offset to place constraints on the body pose of the person and the relative position of the object in 3D space. As shown in Figure \ref{fig:main_pipeline}, our method can be divided into three steps: 1) human-object spatial relation encoding, 2) posterior distribution inference, and 3) post-optimization. In the first step, we construct latent spatial relation space to get a vectorized representation for human-object spatial relation, which is shown in Sec. \ref{section:encoding}. In Sec. \ref{section:distribution}, we present how to infer a coarse distribution for all possible 3D human-object relative arrangements by using normalizing flow. During the optimization stage, we attempt to get a harmonious result that is both aligned well with the image by minimizing the 2D-3D reprojection loss and coherent with posterior knowledge by maximizing the likelihood of potential spatial relation. The details for this optimization process are shown in Sec. \ref{section:optimization}.

\subsection{Spatial Relation Encoding with Human-Object Offset}
\label{section:encoding}
\paragraph{Human-object interaction instance.} To study how human interacts with object in 3D space, we consider the human and the object as a whole and treat this human-object pair as the minimal atomic unit, which is named as human-object interaction instance (HOI instance). For a given human-object pair, there is a trivial way to model it using three components: 1) human mesh modeled by a parametric human body model SMPL \cite{smpl} which defines a mapping $\mathcal{M}(\boldsymbol{\theta}, \boldsymbol{\beta})$ from pose parameters $\boldsymbol{\theta}$ and shape parameters $\boldsymbol{\beta}$ to a body mesh $\boldsymbol{M}_{\text{SMPL}}\in \mathbb{R}^{6890\times 3}$, 2) a pre-scanned object mesh template $\boldsymbol{M}_{\text{object}}$ for target object category, 3) spatial arrangement which is parameterized by the relative translation $\mathbf{t}$ and rotation $\boldsymbol{R}$ of object mesh with respect to the root joint of SMPL. We assume the SMPL is rooted at the origin with zero translation and identity rotation since we actually do not care about the global orientation and translation of SMPL mesh in the context of human-object spatial relation encoding. In this representation, an HOI instance is parameterized using human shape $\boldsymbol{\beta}$, human pose $\boldsymbol{\theta}$, object relative translation $\mathbf{t}$ and object relative rotation $\boldsymbol{R}$. Since the human and the object are treated separately, the relation between the human and the object in 3D space cannot be captured clearly using only relative translation and rotation between human mesh and object mesh. Based on this observation, we propose to use the dense offsets between anchors in human mesh and object mesh to capture a highly detailed correlation between human parts and object parts.

\paragraph{Human-object offset vector.} Since humans can interact with objects in different ways, there will be quite diverse human-object spatial relation patterns. To cover all possible interaction types, we design a simple but general way to encode this. First, we randomly sample $m$ points from the surface of a human mesh to form human anchor set $\mathcal{A}_{\text{SMPL}}$ and $n$ points from the surface of object mesh to form object anchor set $\mathcal{A}_{\text{object}}$. These anchors are sampled only once and we keep them fixed across all human-object interaction instances. Given a human-object interaction instance, the offsets between human anchors and object anchors are calculated by
\begin{equation}
	\mathbf{d}_{i,j} = \mathbf{p}_{j}^\text{o} - \mathbf{p}_i^\text{h}, \mathbf{p}_j^\text{o} \in \mathcal{A}_{\text{object}}, \mathbf{p}_i^\text{h} \in \mathcal{A}_{\text{SMPL}},
	\label{eq:rel_dist}
\end{equation}
where $\mathbf{p}_i^{\text{h}}$ is $i$-th anchor in SMPL mesh $M_{\text{SMPL}}$ and $\mathbf{p}_j^{\text{o}}$ is the $j$-th anchor in object mesh $M_{\text{object}}$. 
We connect all anchors in human anchor set with all anchors in object anchor set to get $m\times n$ offsets. These offsets are concatenated together to form human-object offset vector $\mathbf{x} = (\mathbf{d}_{i,j}) \in \mathbb{R}^{3mn}$. This spatial relationship between the human and the object is encoded within the offset vector. 

\paragraph{Latent spatial relation space construction.} To obtain a more compact representation for human-object spatial relation, auto-encoder \cite{Hinton2006ReducingTD} has been used. However, principal components analysis (PCA) \cite{pca} is more adequate in some cases due to its simplicity and efficiency. Based on these considerations, we use PCA to construct latent spatial relation space. We first collect all human-object instances from the training dataset and calculate offsets between anchors using Eq. (\ref{eq:rel_dist}). For each HOI instance, offsets are concatenated together to form an offset vector $\mathbf{x}$. If there are $t$ HOI instances in the training dataset, we will get a matrix $\boldsymbol{X}\in \mathbb{R}^{t\times3mn}$. PCA is then applied to this matrix to extract the top $k$ component vectors which are mutually orthogonal. These main component vectors form the basis for the latent spatial relation space. Given $\mathbf{x}$, we can project it onto this latent space by linear projection, i.e.
\begin{equation}
	\boldsymbol{\gamma} = \boldsymbol{V}^\text{T}(\mathbf{x} - \boldsymbol{\mu}),
	\label{eq:projection}
\end{equation}
where $\boldsymbol{V} \in \mathbb{R}^{3mn \times k}$ is the projection matrix composed by these component vectors, $\boldsymbol{\mu}$ is the mean vector for offset vector $\mathbf{x}$, and $\gamma$ is a latent vector in this latent spatial relation space. Inversely, we can reproject an arbitrary sample $\boldsymbol{\gamma}$ from latent space $\mathbb{R}^k$ to offset space $\mathbb{R}^{3mn}$ as follows,
\begin{equation}
	\hat{\mathbf{x}} = \boldsymbol{V}\boldsymbol{\gamma} + \boldsymbol{\mu}.
	\label{eq:reprojection}
\end{equation}
By constructing latent spatial relation space in this way, compactness and continuity can be satisfied because of linear projection. Another benefit is that the latent space can be constructed efficiently using PCA technique and there is no need to train a complex neural network.

\paragraph{Recover HOI instance from HO-offset.} An important characteristic of a good representation is that original information can be recovered from it. Human-object offset vector encodes not only spatial arrangement between the human and the object, but also provides constraints on human pose which indicates that we can recover human body pose and object 6D pose from dense human-object offsets by controlling the position of anchors in the surface of the human and the object mesh. Given an arbitrary sample from latent spatial relation space, the offset vectors $\hat{\mathbf{x}}$ can be recovered according to Eq. (\ref{eq:reprojection}). Variables $\{\boldsymbol{\beta}, \boldsymbol{\theta}, \boldsymbol{R}, \mathbf{t}\}$ are calculated from this offset vector approximately by minimizing the $\ell_2$ norm between target offset $\hat{\mathbf{d}}_{i,j}$ and actual offset from the $i$-th anchor of human $p_{i}^\text{h}(\boldsymbol{\theta}, \boldsymbol{\beta})$ to the $j$-th anchor object $p_j^\text{o}(\boldsymbol{R}, \mathbf{t})$, i.e.
\begin{equation}
	\mathcal{L}_{\text{HO-offset}}(\boldsymbol{\theta}, \boldsymbol{\beta}, \boldsymbol{R}, \mathbf{t}) = \sum_{i} \sum_{j} \| p_i^\text{h}(\boldsymbol{\theta}, \boldsymbol{\beta}) + \hat{\mathbf{d}}_{i,j} - p_{j}^\text{o}(\boldsymbol{R}, \mathbf{t}) \|^2.
	\label{eq:rel_loss}
\end{equation}
Note that the positions of human anchor points are controlled by human shape $\boldsymbol{\beta}$ and human pose $\boldsymbol{\theta}$, since they are sampled from SMPL mesh and the position of object anchor points are calculated by:
\begin{equation}
	p_j^\text{o}(\boldsymbol{R}, \mathbf{t}) = \boldsymbol{R} \hat{\mathbf{p}}_j^\text{o} + \mathbf{t},
\end{equation}
where $\hat{\mathbf{p}}_j^o$ is the $j$-th anchor points in object template mesh. Directly optimizing Eq. (\ref{eq:rel_loss}) needs many iterations and may be stuck in local minimum points. The optimization steps can be greatly reduced if we initialize $\{\boldsymbol{\beta}, \boldsymbol{\theta}, \boldsymbol{R}, \mathbf{t}\}$ properly. We first use the neural network to predict human shape $\boldsymbol{\beta}_{\text{init}}$ and human pose $\boldsymbol{\theta}_{\text{init}}$ as described in Sec. \ref{section:distribution} and then substitute $\boldsymbol{\beta}_{\text{init}}$ and $\boldsymbol{\theta}_{\text{init}}$ into (\ref{eq:rel_loss}) to obtain the initial value of $\boldsymbol{R}_{\text{init}}$ and $\mathbf{t}_{\text{init}}$, i.e., solving the following optimization problem:
\begin{equation}
	\boldsymbol{R}_{\text{init}}, \mathbf{t}_{\text{init}} = \mathop{\arg \min}\limits_{\boldsymbol{R},\mathbf{t}} \mathcal{L}_{\text{HO-offset}}(\boldsymbol{\theta}_{\text{init}}, \boldsymbol{\beta}_{\text{init}}, \mathbf{R}, \mathbf{t}).
	\label{eq:rt_init}
\end{equation}
Note that Eq. (\ref{eq:rt_init}) admits a closed-form solution as described in \cite{Choy_2020_CVPR}. 

\subsection{Posterior Distribution Inference by Stacked Normalizing Flow}
\label{section:distribution}

Given an image $I$, we attempt to recover the spatial arrangement of the human-object pair which is encoded using HO-offset. Reconstructing human-object interaction instances from a single-view image is ambiguous due to self-occlusion and mutual-occlusion. Instead of regressing latent human-object spatial arrangement features from the image directly, we follow previous work \cite{Kolotouros_2021_ICCV:prohmr} to model it as probabilistic distribution inference. This distribution inference process requires us to predict the conditional probability $p_{\Gamma|I}(\boldsymbol{\gamma}|\mathbf{c})$ using a bijective function $f_{\text{offset}}$, which transforms a random variable $\mathbf{z}_{\gamma}$ sampled from normal distribution to latent spatial relation feature $\boldsymbol{\gamma}$ with the input image $I$ as condition, i.e.
\begin{equation}
	\boldsymbol{\gamma} = f_{\text{offset}}(\mathbf{z}_{\gamma}|\mathbf{c}),
\end{equation}
where $\mathbf{c}$ is visual feature extracted from input image $I$ using CNN encoder. However, we find that it is not easy to learn this distribution from images directly in practice. To ease the training process, we decouple it into two stacked conditional probabilities:
\begin{equation}
	p_{\Gamma|I}(\boldsymbol{\gamma}|\mathbf{c}) = \int_{\boldsymbol{\theta}} p_{\Gamma|I,\Theta}(\boldsymbol{\gamma}|\mathbf{c},\boldsymbol{\theta})p_{\Theta|I}(\boldsymbol{\theta}|\mathbf{c})\text{d}\boldsymbol{\theta}.
	\label{eq:offset_dist}
\end{equation}
We model it using two different flows: (1) human pose flow conditioned on the input image, (2) offset flow conditioned on the human pose and input image, i.e.
\begin{equation}
	\boldsymbol{\theta} = f_{\text{SMPL}}(\mathbf{z}_{\theta}|\mathbf{c}), \mathbf{z}_{\theta} \sim N(0,I),
	\label{eq:pose_flow}
\end{equation}
and
\begin{equation}
	\boldsymbol{\gamma} = f_{\text{offset}}(\mathbf{z}_{\gamma}|\mathbf{c}, \boldsymbol{\theta}), \mathbf{z}_\gamma \sim N(0,I).
	\label{eq:offset_flow}
\end{equation}

The structure of these stacked normalizing flows is depicted in Figure \ref{fig:main_pipeline}(b). Given an input image, the CNN is used to extract visual feature $\mathbf{c}$ from the image $I$. The initial human shape $\boldsymbol{\beta}$ and the translation of camera $\boldsymbol{T}_{\text{cam}}$ are predicted from $\mathbf{c}$. To infer the posterior distribution of $\boldsymbol{\gamma}$ after observing image $I$, StackFLOW is employed. StackFLOW contains two normalizing flows: human pose flow and offset flow. As formulated in Eq. (\ref{eq:pose_flow}) and Eq. (\ref{eq:offset_flow}), the human pose flow takes visual feature $\mathbf{c}$ as condition to transform a random variable $\mathbf{z}_{\theta}$ sampled from normal distribution to human pose distribution $p_{\Theta|I}(\boldsymbol{\theta}|\mathbf{c})$. We take $\boldsymbol{\theta}_{\text{init}} = \arg \max_{\boldsymbol{\theta}} p_{\Theta|I}(\boldsymbol{\theta}|\mathbf{c})$ as initial value for human pose. Human pose $\boldsymbol{\theta}$ is combined with visual feature $\mathbf{c}$ as the conditions for offset flow to transform random variable $\mathbf{z}_\gamma$ to distribution $p_{\Gamma|I,\Theta}(\boldsymbol{\gamma}|\mathbf{c},\boldsymbol{\theta})$. Combining these two distributions, we can get the posterior distribution $p_{\Gamma|I}(\boldsymbol{\gamma}|\mathbf{c})$ according to Eq. (\ref{eq:offset_dist}).

To train these two normalizing flows, we optimize the network by minimizing the negative log-likelihood of ground-truth $\boldsymbol{\theta}_{\text{gt}}$ and $\boldsymbol{\gamma}_{\text{gt}}$, i.e. the loss function is
\begin{equation}
	\mathcal{L}_{\text{NLL}} = - \ln p_{\Gamma|I;\Theta}(\boldsymbol{\gamma}_{\text{gt}}|\mathbf{c},\boldsymbol{\theta}_{\text{gt}}) - \ln p_{\Theta|I}(\boldsymbol{\theta}_{\text{gt}}|\mathbf{c}).
\end{equation}

In addition to the loss $\mathcal{L}_{\text{SMPL}}$ for supervising SMPL parameters shown in \cite{Kolotouros_2021_ICCV:prohmr}, we introduce another loss for spatial relation feature $\boldsymbol{\gamma}$:
\begin{equation}
	\mathcal{L}_{\gamma} = \lambda_{exp} \mathbb{E}_{\boldsymbol{\gamma}\sim p_{\Gamma|I}} [\| \boldsymbol{\gamma} - \boldsymbol{\gamma}_{\text{gt}} \|_1] + \| \boldsymbol{\gamma}^\star - \boldsymbol{\gamma}_{\text{gt}} \|_1,
\end{equation}
where $\boldsymbol{\gamma}^\star = \arg\max_{\boldsymbol{\gamma}} p_{\Gamma|I}(\boldsymbol{\gamma}|\boldsymbol{c})$. The total training loss is
\begin{equation}
	\mathcal{L}_{\text{train}} = \lambda_{\text{SMPL}} \mathcal{L}_{\text{SMPL}} + \lambda_{\text{NLL}} \mathcal{L}_{\text{NLL}} + \lambda_{\gamma} \mathcal{L}_{\gamma}.
\end{equation}

\subsection{Joint Optimization with Reprojection and Human-Object Offset }
\label{section:optimization}

During inference, we begin with $\mathbf{z}_{\theta} = \boldsymbol{0}, \mathbf{z}_{\gamma} = \boldsymbol{0}$ to get initial human pose $\boldsymbol{\theta}_{\text{init}}$ and latent relation feature $\boldsymbol{\gamma}^\star$. We then use Eq. (\ref{eq:reprojection}) to project latent spatial relation feature $\boldsymbol{\gamma}^\star$ back to offset vector $\mathbf{x}$. The offset $\mathbf{d}_{i,j}$ can be obtained from $\mathbf{x}$ by taking corresponding elements. Finally, we obtain the initial prediction $\{\boldsymbol{\theta}_{\text{init}}, \boldsymbol{\beta}_{\text{init}}, \boldsymbol{R}_{\text{init}}, \mathbf{t}_{\text{init}}\}$ from Eq. (\ref{eq:rt_init}). This initial prediction is based on the distribution with the most likelihood. To make results aligned well with the input image, we need to finetune results with 2D-3D reprojection loss.

Let $\mathbf{J}^{\text{3D}}\in \mathbb{R}^{K \times 3}$ be the 3D joints of human body and $\hat{\mathbf{J}}^{\text{2D}} \in \mathbb{R}^{K\times 2}$ be 2D locations of corresponding joints which are extracted using OpenPose \cite{openpose}, then the 2D-3D reprojection loss for human is defined as
\begin{equation}
	\mathcal{L}_{\text{J}} = \sum_{i=1}^K \| \Pi(\mathbf{J}_i^{\text{3D}}) - \hat{\mathbf{J}}_i^{\text{2D}} \|_1,
	\label{eq:reproj_human}
\end{equation}
where $\Pi: \mathbb{R}^3 \to \mathbb{R}^2$ is the camera projection function.  As for object, we use EPro-PnP \cite{Chen_2022_CVPR:epro_pnp} to get 3D object coordinates $\mathbf{x}^{\text{3D}}\in \mathbb{R}^{N\times 3}$, 2D image coordinates $\mathbf{x}^{\text{2D}} \in \mathbb{R}^{N\times 2}$ and 2D weights $\mathbf{w}^{\text{2D}} \in \mathbb{R}^{N\times 2}_+$, then the 2D-3D reprojection loss for object is defined as
\begin{equation}
	\mathcal{L}_{\text{coor}} = \sum_{i=1}^N \| \mathbf{w}_i^{\text{2D}} \circ (\Pi(\mathbf{R} \mathbf{x}_i^{\text{3D}} + \mathbf{t}) - \mathbf{x}_i^{\text{2D}}) \|_1.
\end{equation}
We also place constraints on human 3D body pose by maximizing the posteriori probabilities of the human body pose $\mathcal{L}_{\text{posteriori}}^{\theta} = - \|\mathbf{z}_{\theta} \|^2$. The loss for 2D-3D reprojection loss is defined as
\begin{equation}
	\mathcal{L}_{\text{2D-3D}} = \lambda_{\text{J}}\mathcal{L}_{\text{J}} + \lambda_{\text{coor}} \mathcal{L}_{\text{coor}} + \lambda_{\text{posteriori}}^{\theta} \mathcal{L}_{\text{posteriori}}^\theta.
\end{equation}
The 2D-3D reprojection loss aims at aligning results with image content without considering the interaction between the human and the object. To restrict the relative offset between the human and the object, we add the offset loss $\mathcal{L}_{\text{HO-offset}}$ shown in Eq. (\ref{eq:rel_loss}) and posteriori distribution loss $\mathcal{L}_{\text{posteriori}}^\gamma = - \|\mathbf{z}_{\gamma}\|^2$, which form the loss for human-object spatial relation
\begin{equation}
	\mathcal{L}_{\text{offset}} = \lambda_{\text{HO-offset}} \mathcal{L}_{\text{HO-offset}} + \lambda_{\text{posteriori}}^\gamma \mathcal{L}_{\text{posteriori}}^\gamma.
\end{equation}
Finally, optimization loss is defined as
\begin{equation}
	\mathcal{L}_{\text{optim}} = \mathcal{L}_{\text{2D-3D}} + \mathcal{L}_{\text{offset}}.
	\label{eq:optim_losses}
\end{equation}
	
	\section{Experiments}
	\begin{figure*}[htbp]
	\centering
	\includegraphics[width=\linewidth]{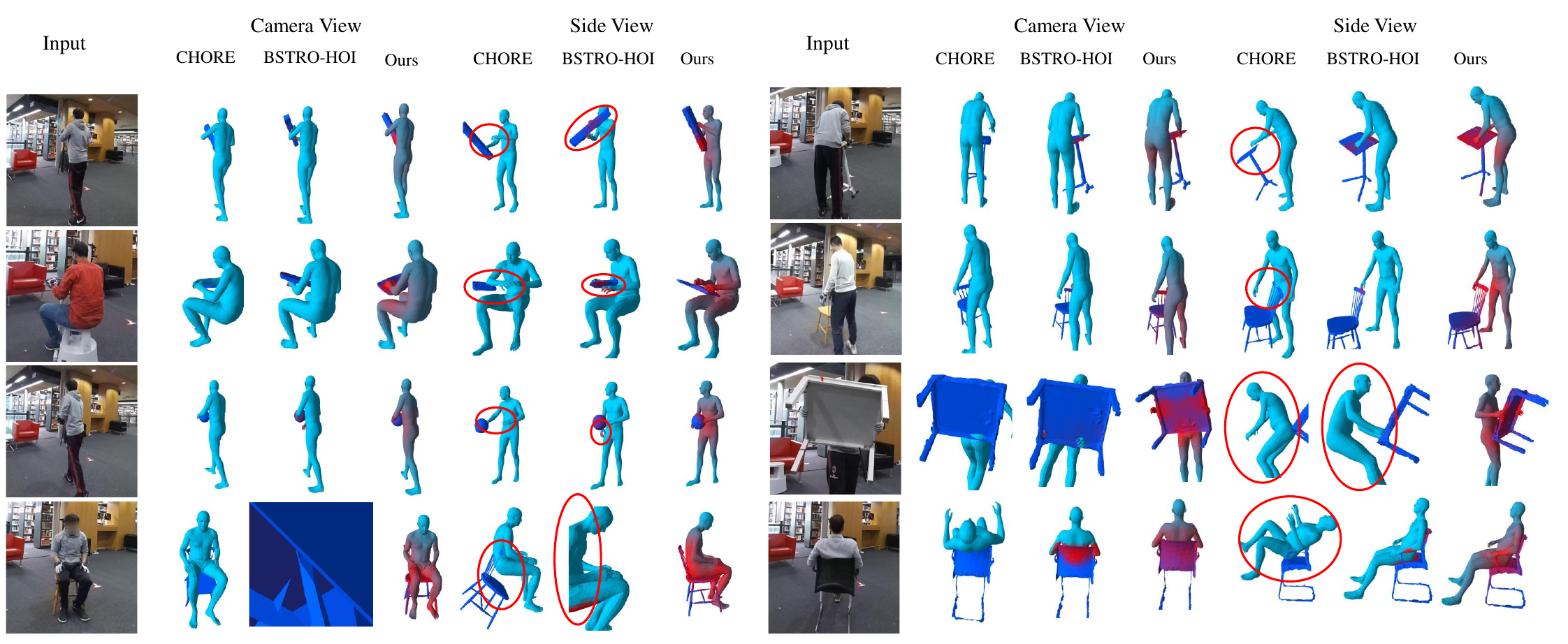}
	\caption{Visualized reconstruction results on BEHAVE dataset. The red regions depict the contact region in BSTRO or the relative distance of our method. The red circles mark the incorrect reconstruction results. These results show that our method performs well in some heavy occlusion cases.}
	\label{fig:comparison_occlusion}
\end{figure*}
	
\paragraph{Dataset.} We conduct experiments on two indoor datasets: BEHAVE \cite{Bhatnagar_2022_CVPR:BEHAVE} and InterCap \cite{intercap}. BEHAVE is a recently released dataset that captures 8 subjects interacting with 20 different objects indoors using a multi-view camera capture system. We follow the official train/test split to train and test our method. Due to the cost of collecting annotations, BEHAVE doesn't provide enough training data which will easily cause the overfitting problem. To prevent this, we render fake images with new viewpoints and new subjects to augment the original training dataset.  InterCap is a larger dataset which contains 4M images of 10 subjects interacting with 10 objects. 
% Like BEHAVE, the InterCap dataset also provides pseudo labels for SMPL and object 6D pose. Since the official dataset doesn't provide the train/test split, we have to split the dataset by ourselves. 
We randomly select 20\% sequences for testing and the rest for training which results in 326, 955 images in the training split and 73, 541 images in the testing split.

\paragraph{Free-viewport augmentation.} We apply free-viewport data augmentation to generate new images. For each HOI instance sampled from the training dataset, we first use MetaAvatar \cite{meta_avatar} trained on CAPE dataset \cite{CAPE:CVPR:20,Pons-Moll:Siggraph2017} to generate clothed human mesh given human pose $\boldsymbol{\theta}$ and place it with object mesh template, which is transformed by $\boldsymbol{R}$ and $\mathbf{t}$, in world coordinate. We then render new images by changing the viewport of the camera to simulate all possible occlusions between the human and the object. In the end, we render 12 images with different viewports for each HOI instance in the training dataset. These rendered fake images are used as a supplementary dataset to train our model. 
% More details and visualization can be found in the supplementary material.

\paragraph{Evaluation metric.} Following previous works \cite{Bhatnagar_2022_CVPR:BEHAVE}, we use Chamfer distance to evaluate the quality of the reconstructed mesh. For a fair comparison, we assume the object label and bounding box are known before, what we need to predict are the SMPL parameters and the object's 6D pose. With the reconstructed SMPL mesh and object mesh, we first align them with ground truth meshes using Procrustes analysis, then Chamfer distance is calculated based on the point clouds sampled from reconstructed meshes and ground truth meshes. 

\iffalse
\paragraph{Implementation details.} We sample 64 anchors from the surface of the object template mesh and 704 anchors from the surface of the SMPL template mesh. The dimension of relative distance vector $\mathbf{x}$ is reduced from $704\times 64 \times 3$ to 32. We use ResNet-50 as CNN backbone and use Glow with 4 layers as distance flow. On BEHAVE dataset, our model train 120 epochs with a learning rate of 1e-4 and another 8 epochs with a learning rate of 1e-5. We use Adam optimizer and the batch size is set to 32. Our model can be trained in one GPUs within two days. 
\fi

\subsection{Comparisons with the State-Of-The-Arts}
We compare our method with three state-of-the-art methods: PHOSA \cite{zhang2020phosa}, CHORE \cite{xie2022chore} and BSTRO \cite{Huang_2022_CVPR:BSTRO} on BEHAVE and InterCap dataset. PHOSA is an optimization-based framework that targets at reconstructing human-object spatial arrangement from image in the wild. CHORE is a learning-based method that learns to jointly reconstruct the human and the object from a single RGB image. BSTRO is a powerful model which predicts human-scene contact from a single image. To compare with contact-based models, we adapt it to the task of human-object reconstruction. We name this baseline as BSTRO-HOI. More details about BSTRO-HOI can be found in supplementary Materials.
\begin{table}[htbp]
	\centering
	\footnotesize
	\tabcolsep=1.5pt
	\resizebox{\linewidth}{!}{
		\begin{tabular}{ @{}lcccc@{} }
			\toprule
			& \multicolumn{2}{c}{BEHAVE}  & \multicolumn{2}{c}{InterCap}\\
			\cmidrule{2-5}
			Method & SMPL $\downarrow$ & Object $\downarrow$ & SMPL $\downarrow$ & Object $\downarrow$ \\
			\midrule
			PHOSA & 12.17 $\pm$ 11.13 & 26.62 $\pm$ 21.87 & 6.06 $\pm$ 11.13 & 14.81 $\pm$ 11.96\\
			CHORE & 5.58 $\pm$ 2.11 & 10.66 $\pm$ 7.71 & 6.86 $\pm$ 2.45 & 15.49 $\pm$ 10.13\\
			BSTRO-HOI  & 4.77 $\pm$ 2.46 & 11.08 $\pm$ 13.14 & 4.80 $\pm$ 2.82 & 9.70 $\pm$ 11.05\\
			\midrule
			Ours & 4.61 $\pm$ 2.04 & 9.86 $\pm$ 9.59 & \textbf{4.42 $\pm$ 1.85} & \textbf{8.04 $\pm$ 7.37}\\
			Ours$\dagger$ & \textbf{4.33 $\pm$ 1.83} & \textbf{8.87 $\pm$ 8.76} & - & - \\
			\bottomrule
		\end{tabular}
	}
	\caption{Comparison of mean and standard deviation of Chamfer distance (cm) over all HOI instances on BEHAVE and InterCap datasets. $\dagger$ indicates the model is trained with augmented dataset. \textbf{Blod} indicates the best result.}
	\label{tab:main_table}
\end{table}

\paragraph{Quantitative evaluation.} As shown in Table \ref{tab:main_table}, we compare our method with baseline methods on BEHAVE dataset and InterCap dataset. Our method achieves competitive results compared with state-of-the-art methods. Compared with pure optimization-based method PHOSA, all learning-based methods show incomparable advantages. Compared with other learning-based methods, our method achieves more accurate results, which indicates that human-object offset is a more suitable representation to encode human-object spatial relation.

\paragraph{Qualitative evaluation.} We also compare our method against CHORE and BSTRO-HOI qualitatively for heavy occlusion cases in Figure \ref{fig:comparison_occlusion}. From these cases, we can see that when objects are heavily occluded by human or some human body parts are heavily occluded by object, our method can still draw hints from visible human body parts or objects to guess the potential position of object or potential human body pose by means of HO-offset. As BSTRO-HOI depends on good initilization of human pose and object pose, it fails on the cases in which the object or the human is almost unseen. CHORE also has the same problem. On the contrary, our method is more robust on these heavy occlusion cases. 
% On the contrary, since the initial human pose and object pose are inferred by ourselves, our method is more robust on these heavy occlusion cases. 

\paragraph{Method complexity comparison.} We compare different methods in terms of space efficiency and time efficiency in Table \ref{tab:main_efficiency}. Our method makes a good balance between space complexity and computation complexity. It is noteworthy that our method outperforms CHORE from 7.90 to 6.60 (with 16\% improvement) in terms of reconstruction accuracy with a dramatic reduction from 366.04 to 43.39 (with 88\% reduction) during the optimization stage. This dramatic reduction of time consumed in the post-optimization stage benefits from two aspects. First, before post-optimization, we have already got a good initialization which is predicted by StackFLOW, only a few iterations are needed to get the optimal results. The other factor that contributes to dramatic time reduction is the simplicity and efficiency of our optimization loss terms. 
On the contrary, CHORE relies on multi-stage optimization and complex losses for CHORE field fitting to get accurate reconstruction results.

\begin{table}[htbp]
	\centering
	\footnotesize
	\tabcolsep=1.5pt
	\resizebox{\linewidth}{!}{
	\begin{tabular}{ @{}lcccc@{} }
		\toprule
		Method & \#Params (M) & GFLOPs & Time (s) & Chamfer Dist.  \\
		\midrule
		PHOSA & - & - & 14.23 & 19.40 \\
		CHORE & \textbf{18.19} & 396.39 & 366.04 & 7.90 \\
		BSTRO-HOI & 146.99 & 40.20 & 18.90 & 7.40 \\
		\midrule
		Ours (w/o optim.) & 77.02 & \textbf{5.50} & \textbf{1.15} & 9.34 \\
		Ours (w optim.) & 77.02 & \textbf{5.50} & 43.39 & \textbf{6.60}  \\
		\bottomrule
	\end{tabular}
	}
	\caption{Time and space complexity comparisons on BEHAVE dataset. The second and third columns compare the size and computation of neural network during inference. The fourth column compares the time spent processing each image. The time is tested on a single NVIDIA GeForce RTX 2080 Ti GPU. The last column compares the reconstruction error of different methods. The chamfer distance is averaged between SMPL and object.}
	\label{tab:main_efficiency}
\end{table}

\subsection{Ablation Study}
\paragraph{Effectiveness of offset loss.} To demonstrate the effectiveness of offset loss in the stage of post-optimization, we report the results with and without the offset loss in Table \ref{tab:ablation_rel_dist_loss}. Without any optimization, our method can already achieve comparable performance. If we optimize only with reprojection loss, the accuracy of reconstruction becomes worst due to the incorrect of coordinate map predicted by EPro-PnP \cite{Chen_2022_CVPR:epro_pnp}. Only if we jointly optimize with offset loss and 2D-3D reprojection loss, the best performance can be achieved. 
 % The offset loss places constraints on the spatial relation between human and object and suppress the wrong pose of object. With 2D-3D reprojection loss, the initial prediction of our neural network can be aligned better with input image.

\begin{table}[htbp]
	\centering
	\tabcolsep=1.5pt
	\footnotesize
	\resizebox{\linewidth}{!}{
	\begin{tabular}{ @{}llcccc@{} }
		\toprule
		&& \multicolumn{2}{c}{BEHAVE}  & \multicolumn{2}{c}{InterCap}\\
		\cmidrule{3-6}
		$\mathcal{L}_{\text{offset}}$ & $\mathcal{L}_{\text{2D-3D}}$  & SMPL $\downarrow$ & Object $\downarrow$ & SMPL $\downarrow$ & Object $\downarrow$\\
		\midrule
		& & 4.83 $\pm$ 2.06 & 13.85 $\pm$ 11.88 & 4.96 $\pm$ 2.26 & 11.53 $\pm$ 10.56 \\
		\checkmark & & 5.68 $\pm$ 2.25 & 13.85 $\pm$ 12.17 & 5.75 $\pm$ 2.52 & 12.25 $\pm$ 10.83 \\
		&\checkmark & 4.79 $\pm$ 2.44 & 15.15 $\pm$ 18.19  & 5.71 $\pm$ 3.35 & 17.27 $\pm$ 15.30 \\
		\checkmark & \checkmark & \textbf{4.33 $\pm$ 1.83} & \textbf{8.87 $\pm$ 8.76} & \textbf{4.42 $\pm$ 1.85} & \textbf{8.04 $\pm$ 7.37} \\	
		\bottomrule
	\end{tabular}
	}
	\caption{Effectiveness of different loss during the process of post-optimization. $\mathcal{L}_{\text{offset}}$ denotes the loss about offset loss and $\mathcal{L}_{\text{2D-3D}}$ denotes the loss about 2D-3D reprojection in Eq. (\ref{eq:optim_losses}).}
	\label{tab:ablation_rel_dist_loss}
\end{table}

% \iffalse
\paragraph{Effectiveness of data augmentation.}  In Table \ref{tab:ablation_aug}, we list the performance of different methods trained with augmented dataset or without augmented dataset. After trained along with our augmented dataset, the performance can be improved across all methods. Whether using generated data or not, our method outperforms other state-of-the-art methods.

\begin{table}[htbp]
	\centering
	\footnotesize
	%\normalsize
	\begin{tabular}{ @{}lccc@{} }
		\toprule
		Method & Data AUG. & SMPL $\downarrow$ & Object $\downarrow$ \\
		\midrule
		\multirow{2}{*}{CHORE} & \ding{55} & 5.58 $\pm$ 2.00 & 10.66 $\pm$ 7.71 \\
		& \checkmark & \textbf{5.52 $\pm$ 2.00} & \textbf{10.27 $\pm$ 7.75} \\
		\midrule
		\multirow{2}{*}{BSTRO-HOI} & \ding{55} & 4.77 $\pm$ 2.46 & 11.08 $\pm$ 13.14 \\
		& \checkmark & \textbf{4.50 $\pm$ 2.28} & \textbf{10.29 $\pm$ 12.09} \\
		\midrule
		\multirow{2}{*}{Ours} & \ding{55} & 4.61 $\pm$ 2.04 & 9.86 $\pm$ 9.59 \\
		& \checkmark & \textbf{4.33 $\pm$ 1.83} & \textbf{8.87 $\pm$ 8.76} \\
		\bottomrule
	\end{tabular}
	\caption{Ablation studies on BEHAVE dataset for the effectiveness of data augmentation.}
	\label{tab:ablation_aug}
\end{table}

	\section{Conclusion}
	In this work, we show how to encode and capture highly detailed 3D human-object spatial relations from single-view images using Human-Object Offset. Towards monocular human-object reconstruction, a Stacked Normalizing Flow is proposed to infer the posterior distribution of human-object spatial relation from a single-view image. During the optimization stage, offset loss is proposed to constrain the body pose of humans and the relative 6D pose of objects. Our method outperforms state-of-the-art models on two challenging benchmarks including BEHAVE or InterCap dataset. Especially, our model is good at handling heavy occlusion cases. Even if the objects are heavily occluded by the human, our method can still draw cues from visible human pose to infer the potential pose of the objects. 
% However, some limitations may exist in our method such as the lack of generalization ability and heavy dependence on the diversity of human-object spatial relation in the training set.
	
	\clearpage
	\section*{Acknowledgments}
	This work was supported by the Shanghai Sailing Program (21YF1429400, 22YF1428800), Shanghai Local College Capacity Building Program (23010503100,22010502800), NSFC programs (61976138, 61977047), the National Key Research and Development Program (2018YFB2100500), STCSM (2015F0203-000-06), SHMEC (2019-01-07-00-01-E00003) and Shanghai Frontiers Science Center of Human-centered Artificial Intelligence (ShangHAI).
	
	\bibliographystyle{named}
	\bibliography{ijcai23}
	
\end{document}

\maketitle
	
	\begin{abstract}
		In this supplementary document we provide implementation details for our method and baseline method in Sec. 1. In Sec. 2 we show some samples from our augmented dataset with novel viewports. More qualitative comparisons with baseline methods and quantitative ablation studies are shown in Sec 3.
	\end{abstract}
	
	\section{Implementation Details}
	\subsection{Sampled Anchors}
	\begin{figure}[htbp]
		\centering
		\includegraphics[width=\linewidth]{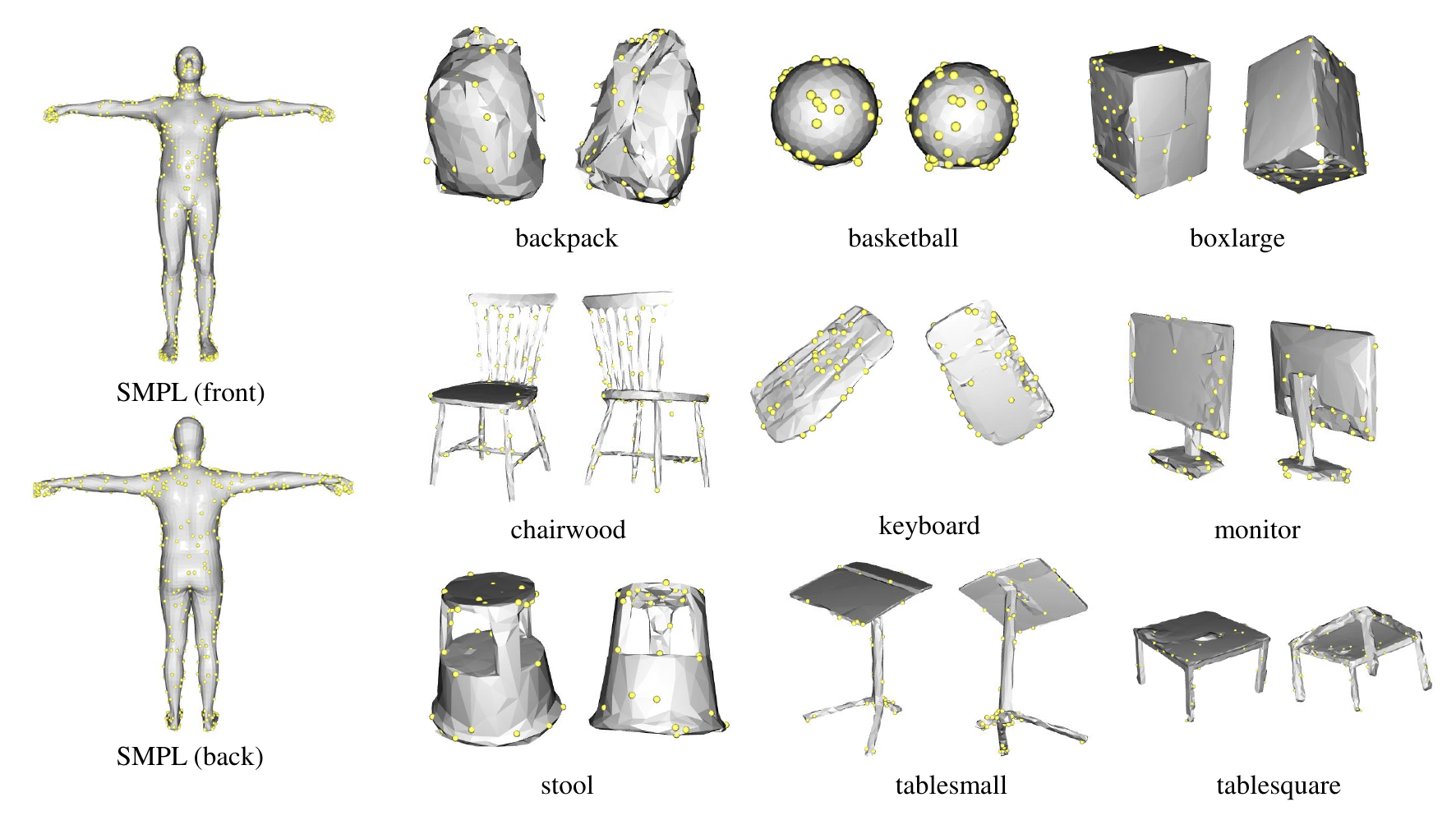}
		\caption{In this figure, we show anchors sampled from the surface of SMPL mesh and object mesh in yellow points.}
		\label{fig:anchors}
	\end{figure}
	In order to get dense offsets between the human and the object, we sample about 64 anchors from the surface of object mesh and 704 ($32\times 22$) anchors from the surface of SMPL mesh as shown in figure \ref{fig:anchors}. We connect all object anchors with SMPL anchors to get $22,528$ directed relative offsets. These offsets are projected into human-object spatial relation latent space which has 32 dimensions.
	
	\subsection{Detailed Network Structure}
	
	\begin{figure}[htbp]
		\centering
		\includegraphics[width=\linewidth]{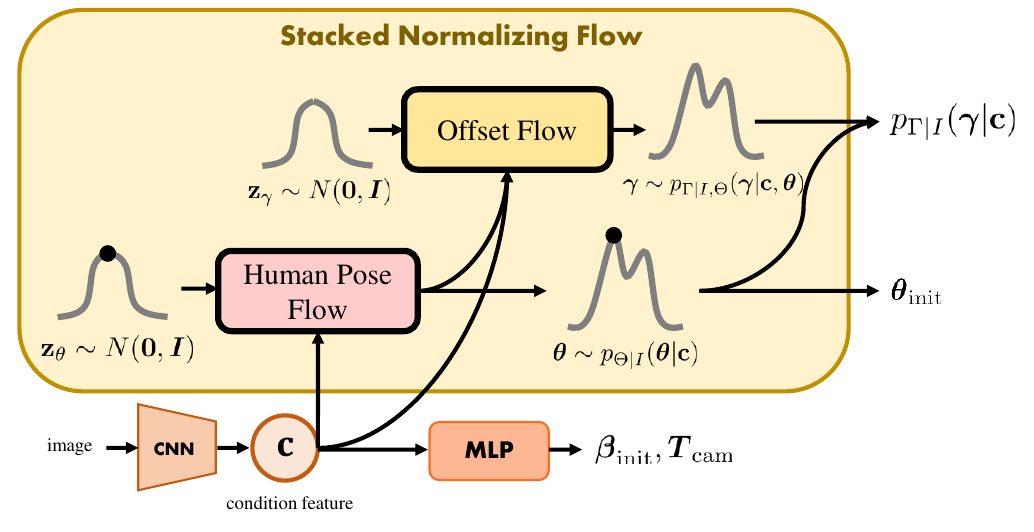}
		\caption{This figure shows the detailed network structure of pose-guided relative distance normalizing flow.}
		\label{fig:distance_flow}
	\end{figure}
	
	Figure \ref{fig:distance_flow} shows the detailed structure of stacked normalizing flow. Given an input image $I$, we first use ResNet-50 to encode it into a condition features $\mathbf{c}\in\mathbb{R}^{2048}$. Multi-linear perceptron network is utilized to predict the shape of person $\boldsymbol{\beta}_{\text{init}}$ and the translation of camera $\boldsymbol{T}_{\text{cam}}$ from this condition features. In order to infer the posteriori distribution of human-object spatial relation $p_{\Gamma|I}(\boldsymbol{\gamma}|\mathbf{c})$, stacked normalizing flow is deployed, which contains two normalizing flows: human pose flow and offset flow. Human pose flow transforms $\mathbf{z}_{\theta}$ which is sampled from normal distribution to the human pose distribution $p_{\Theta|I}(\boldsymbol{\theta}|\mathbf{c})$ conditioned by CNN feature $\mathbf{c}$. We use $\boldsymbol{\theta}_{\text{init}} = \boldsymbol{\theta}^\star = \arg\max_{\boldsymbol{\theta}}p_{\Theta|I}(\boldsymbol{\theta}|\mathbf{c})$ as the initial value of human pose. As for offset flow, it transforms another random variable $\mathbf{z}_{\gamma}\sim N(\boldsymbol{0}, \boldsymbol{I})$ to offset distribution given human pose $\boldsymbol{\theta}$ and CNN feature $\mathbf{c}$ as conditions. Combining two normalizing flows, we can get the distribution for human-object offsets, i.e.
	\begin{equation}
		p_{\Gamma|I}(\boldsymbol{\gamma}|\mathbf{c}) = \int_{\boldsymbol{\theta}} p_{\Theta|I}(\boldsymbol{\theta}|\mathbf{c}) p_{\Gamma|I,\Theta}(\boldsymbol{\gamma}|\mathbf{c},\boldsymbol{\theta}) \text{d}\boldsymbol{\theta}.
	\end{equation}
	To get $\boldsymbol{\gamma}^\star = \arg \max_{\boldsymbol{\gamma}} p_{\Gamma|I}(\boldsymbol{\gamma}|\mathbf{c})$, we use a greedy strategy to get a suboptimal solution
	\begin{equation}
		\boldsymbol{\gamma}^\star \gets \arg \max_{\boldsymbol{\gamma}}p_{\Gamma|I,\Theta} (\boldsymbol{\gamma}|\mathbf{c},\boldsymbol{\theta}^\star).
	\end{equation}
	Using $\boldsymbol{\gamma}^\star$, $\boldsymbol{\theta}_{\text{init}}$ and $\boldsymbol{\beta}_{\text{init}}$, we can get an initial prediction for human-object interaction instance.
	
	\begin{figure*}[htbp]
		\centering
		\includegraphics[width=\linewidth]{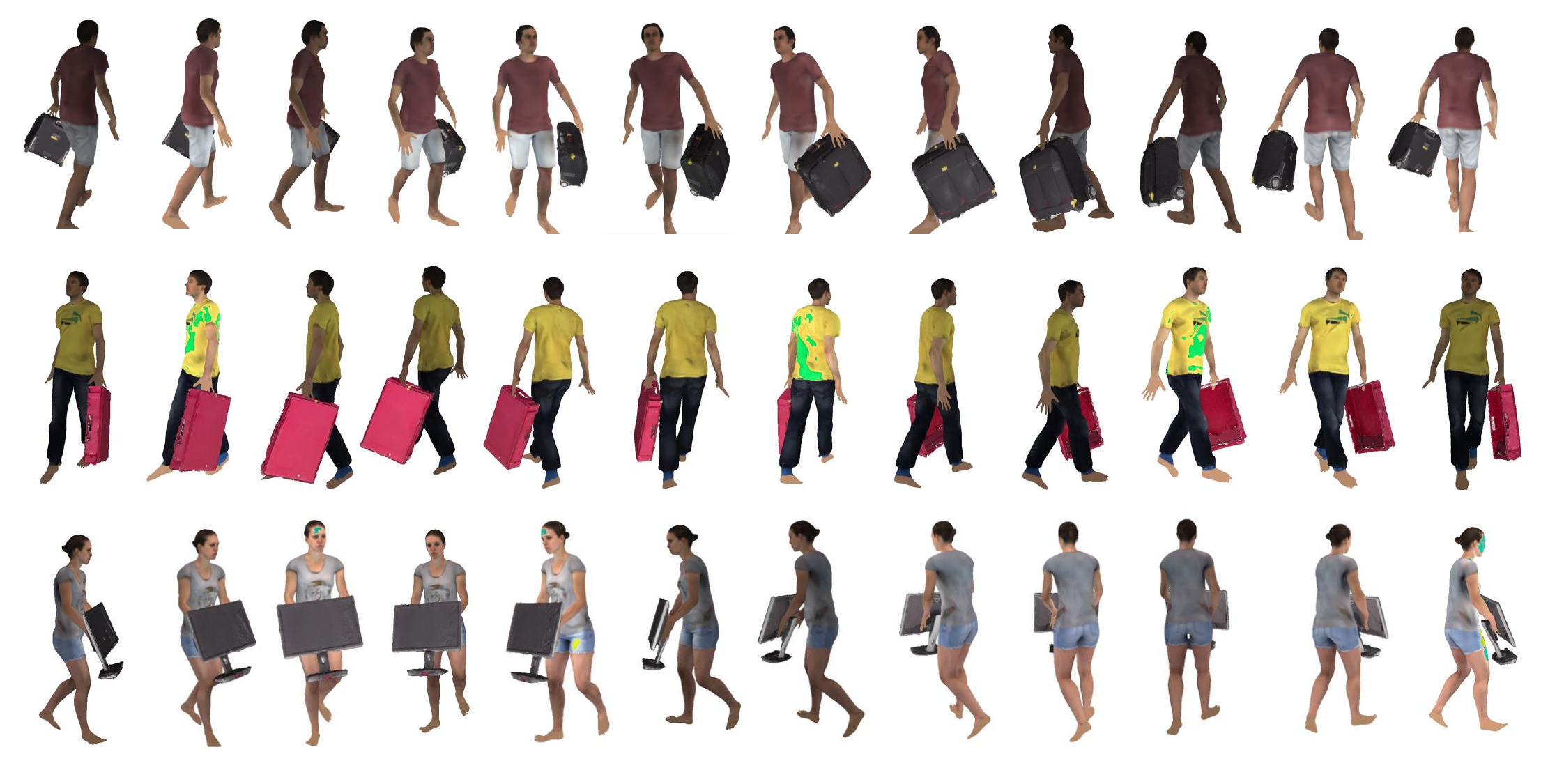}
		\caption{This figure shows different views for each HOI instance.}
		\label{fig:data_aug_view}
	\end{figure*}
	
	\subsection{Other Details}
	We sample 64 anchors from the surface of the object template mesh and 704 anchors from the surface of the SMPL template mesh. The dimension of offset vector $\mathbf{x}$ is reduced from $704\times 64 \times 3$ to 32. We use ResNet-50 as CNN backbone and use Glow with 4 layers as offset flow. On BEHAVE dataset, our model train 120 epochs with a learning rate of 1e-4 and another 8 epochs with a learning rate of 1e-5. We use Adam optimizer and the batch size is set to 32. Our model can be trained in one GPUs within two days. During post-optimization stage, we tune loss weights empirically. In our experiments, we set $\lambda_\text{J},\lambda_{\text{posterior}}^{\theta}$ to $0.1$ and set $\lambda_{\text{posterior}}^{\gamma},\lambda_{\text{HO-offset}}$ to $1$. To adaptively suppress the wrong coordinate maps predicted by Epro-PnP, we calculate $\lambda_{\text{coor}}$ by
	\begin{equation}
		\lambda_{\text{coor}} = e^{\frac{\sum_{i=1}^{N}\sum_{j = 1}^{2}\mathbf{w}_{ij}^{\text{2D}} - k_1}{k_2}},
	\end{equation}
	where $\mathbf{w}_{ij}^{\text{2D}} \in \mathbb{R}^{N\times 2}$ is the weights for each object coordinate, $k_1$ is set to 15 and $k_2$ is set to 10 empirically in BEHAVE dataset.

	\subsection{BSTRO-HOI Implementation Details}
	To compare with the method, which is based on contact map, we build a baseline method which is based on BSTRO. BSTRO is a model that predicts the contact map in SMPL mesh from image. We adapt it into the task of human-object interaction reconstruction. We name this baseline as BSTRO-HOI. In BSTRO-HOI, we introduce two Transformer decoder: SMPL contact Transformer decoder and object contact Transformer decoder. The SMPL contact Transformer decoder remains unchanged with original Transformer decoder. It predicts a score $p^{\text{h}}_i \in [0, 1]$ for each vertex in SMPL mesh, which encodes the probability of $i$-th vertex being in contact. But the object contact Transformer decoder is added newly to predict contact map in object mesh. The object contact Transformer decoder shares the same structure with SMPL contact Transformer decoder, except for different meanings of input and output. The object contact Transformer decoder takes vertices in object mesh template as position encoding and outputs the scores $p^{\text{o}}$ indicating which regions are in contact. We train BSTRO-HOI in BEHAVE dataset for 20 epochs with learning rate of 1e-4. During optimization stage, we use contact map loss to place constraint on human body pose and object relative position which is defined as
	\begin{equation}
		\mathcal{L}_{\text{contact}} = \frac{1}{\sum_{i} p_{i}^{\text{h}}} \sum_{i} p_{i}^{\text{h}} \cdot \text{dist}(\mathbf{v}_i^{\text{h}}, \mathcal{M}_{\text{o}}) + \frac{1}{\sum_{j} p_{j}^{\text{o}}} \sum_{j} p_{j}^{\text{o}} \cdot \text{dist}(\mathbf{v}_j^{\text{o}}, \mathcal{M}_{\text{h}}),
	\end{equation}
	where $\mathcal{M}_{\text{h}}\in\mathbb{R}^{6890\times3}$ is SMPL mesh, $\mathcal{M}_{\text{h}}\in\mathbb{R}^{k\times3}$ is object mesh and the function $\text{dist}(\mathbf{u}, \mathcal{M})$ calculates the minimal distance between vertex $\mathbf{u}$ and mesh $\mathcal{M}$ which is defined as
	\begin{equation}
		\text{dist}(\mathbf{u}, \mathbf{M}) = \min_{\mathbf{v}\in \mathcal{M}} \| \mathbf{u} - \mathbf{v} \|_2.
	\end{equation}
	We optimize contact loss jointly with human joint reprojection loss $\mathcal{L}_{\text{J}}$ and object 2D-3D reprojection loss $\mathcal{L}_{\text{coor}}$. 
	
	\begin{figure}[htbp]
		\centering
		\includegraphics[width=\linewidth]{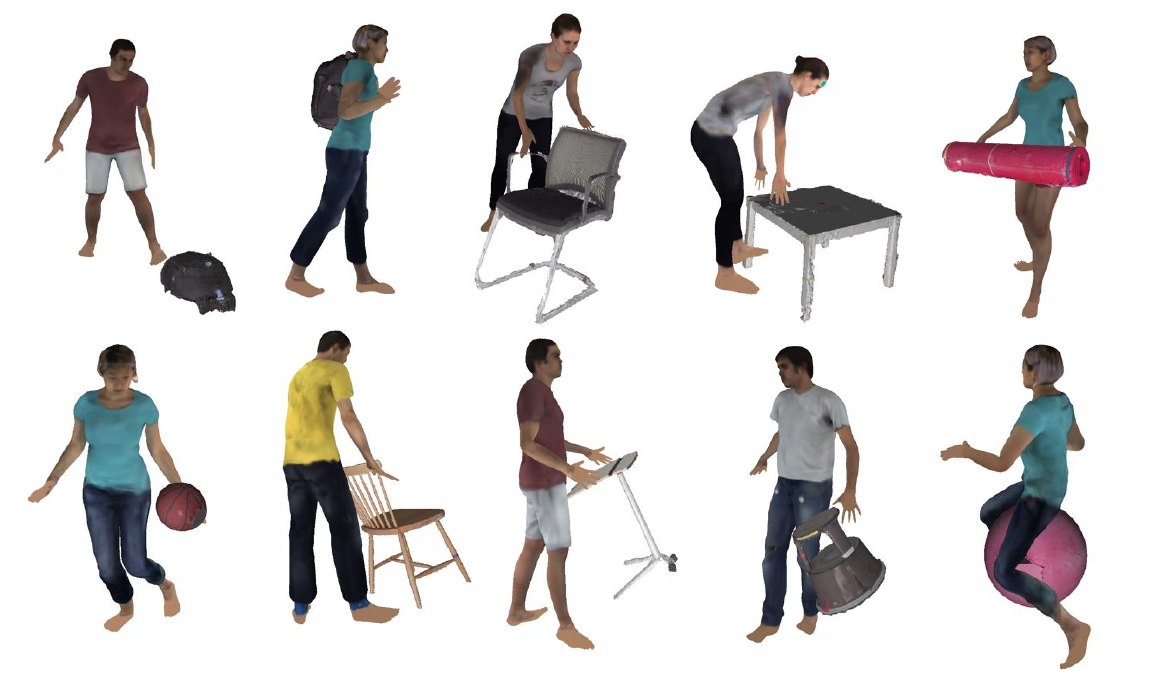}
		\caption{This figure shows the images rendered by our free viewport data augmentation pipeline.}
		\label{fig:data_aug}
	\end{figure}
	\section{Free Viewport Data Augmentation}
	In Figure \ref{fig:data_aug} we show the images generated by our free viewport data augmentation pipeline. Figure \ref{fig:data_aug_view} shows different views for each HOI instance.

	\section{More Experiment Results}
	
	\iffalse
	\subsection{Comparison on Occlusion Cases}
	\begin{table}[htbp]
		\centering
		\footnotesize
		\hspace{-3mm}
		\begin{tabular}{ @{}lcc@{} }
			\toprule
			Method & SMPL $\downarrow$ & Object $\downarrow$  \\
			\midrule
			BSTRO-HOI  & 6.17 $\pm$ 7.70 & 14.95 $\pm$ 17.08 \\
			CHORE & 4.96 $\pm$ 1.72 & 12.84 $\pm$ 9.36 \\
			\midrule
			Ours$\dagger$ & \textbf{4.31 $\pm$ 1.68} & \textbf{10.48 $\pm$ 9.60} \\
			\bottomrule
		\end{tabular}
		\vspace{-2mm}
		\caption{Comparison of mean and standard deviation of Chamfer distance (cm) over all HOI instances on BEHAVE dataset. $\dagger$ indicates the model is trained with augmented dataset. \textbf{Blod} indicates the best result.}
		\label{tab:main_table_behave}
		\vspace{-4mm}
	\end{table}
	\fi

	\subsection{Other Ablation Studies}
	\paragraph{Evaluation on the number of anchors.} The number anchors sampled from the surface of SMPL mesh and object mesh affect the fineness of spatial relationship. The more anchors, the more details for spatial relationship can be captured in relative distances. We study the influence of the number of anchors on the reconstruction accuracy by varying the number of anchors. As shown in Table \ref{tab:albation_num_anchors}, the more anchors we use, the better of performance can be achieved. But taking the computation cost and memory size into consideration, we can not sample too many anchors.  
	
	\begin{table}[htbp]
		\centering
		\footnotesize
		\hspace{-3mm}
		\begin{tabular}{ @{}llcc@{} }
			\toprule
			$n_{\text{SMPL}}$ & $n_{\text{object}}$ & SMPL $\downarrow$ & Object $\downarrow$ \\
			\midrule
			88 & 4 & 4.40 $\pm$ 1.83 & 9.63 $\pm$ 9.33  \\
			176 & 8 & 4.41 $\pm$ 1.88 & 9.24 $\pm$ 8.85 \\
			352 & 16 & 4.34 $\pm$ 1.82 & 8.95 $\pm$ 8.86 \\
			704 & 32 & \textbf{4.31 $\pm$ 1.81} & \textbf{8.85 $\pm$ 8.73} \\
			\bottomrule
		\end{tabular}
		\vspace{-2mm}
		\caption{Ablation studies on BEHAVE dataset over the number of anchors. $n_{\text{SMPL}}$ and $n_{\text{object}}$ are the numbers of anchors sampled from the surface of SMPL mesh and object mesh respectively.}
		\label{tab:albation_num_anchors}
		\vspace{-4mm}
	\end{table}
	
	\paragraph{Evaluation on the fraction of free-viewport augmentation used for training.} We evaluate the models trained with different fraction of augmented dataset in Table \ref{tab:albation_fraction_fakedata}. With full of augmented dataset, the Chamfer distance of object drops about 1 point. We also observe that the improvement is slight if we increase the fraction of augmented dataset from 0.5 to 1, which indicates that half of the augmented dataset is enough for training.
	
	\begin{table}[htbp]
		\centering
		\footnotesize
		\hspace{-3mm}
		\begin{tabular}{ @{}lcc@{} }
			\toprule
			$\alpha$ & SMPL $\downarrow$ & Object $\downarrow$\\
			\midrule
			0 & 4.60 $\pm$ 2.04 & 9.86 $\pm$ 9.59 \\
			0.17 & 4.39 $\pm$ 1.84 & 9.11 $\pm$ 8.67 \\
			0.25 & 4.40 $\pm$ 1.83 & 9.14 $\pm$ 8.73 \\
			0.33 & 4.37 $\pm$ 1.86 & 9.03 $\pm$ 8.98 \\
			0.50 & 4.33 $\pm$ 1.82 & 8.88 $\pm$ 8.53 \\
			1 & \textbf{4.33 $\pm$ 1.83} & \textbf{8.87 $\pm$ 8.76} \\
			\bottomrule
		\end{tabular}
		\vspace{-2mm}
		\caption{Ablation studies on BEHAVE dataset over the fraction of free-viewport augmentation used for training.}
		\label{tab:albation_fraction_fakedata}
		\vspace{-4mm}
	\end{table}
	
	\subsection{More Qualitative Results}
\begin{figure}[htbp]
		\centering
		\includegraphics[width=\linewidth]{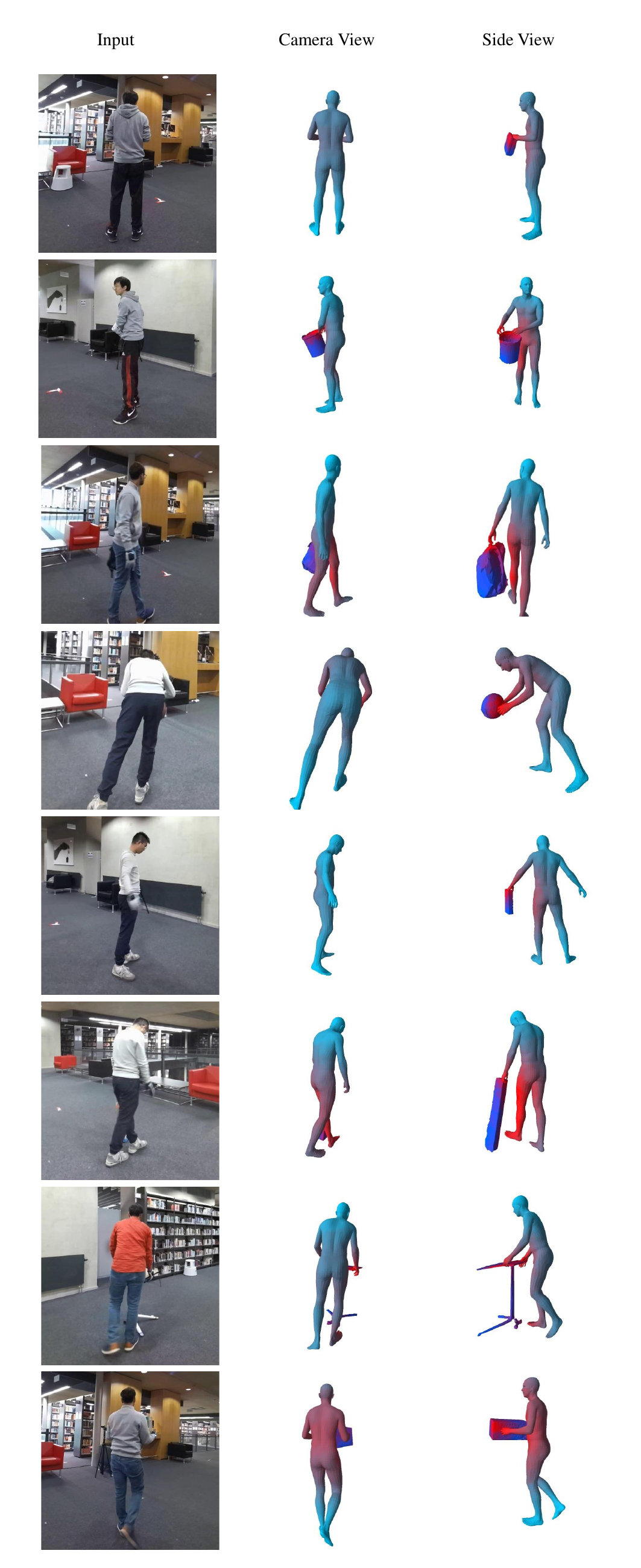}
		\caption{This figure shows heavy occlusion cases. Even if the object is almost unseen, our method can still guess the position of object by drawing hints from visible human.}
		\label{fig:comparison_occlusion}
	\end{figure}

\begin{figure*}[htbp]
		\centering
		\includegraphics[width=\linewidth]{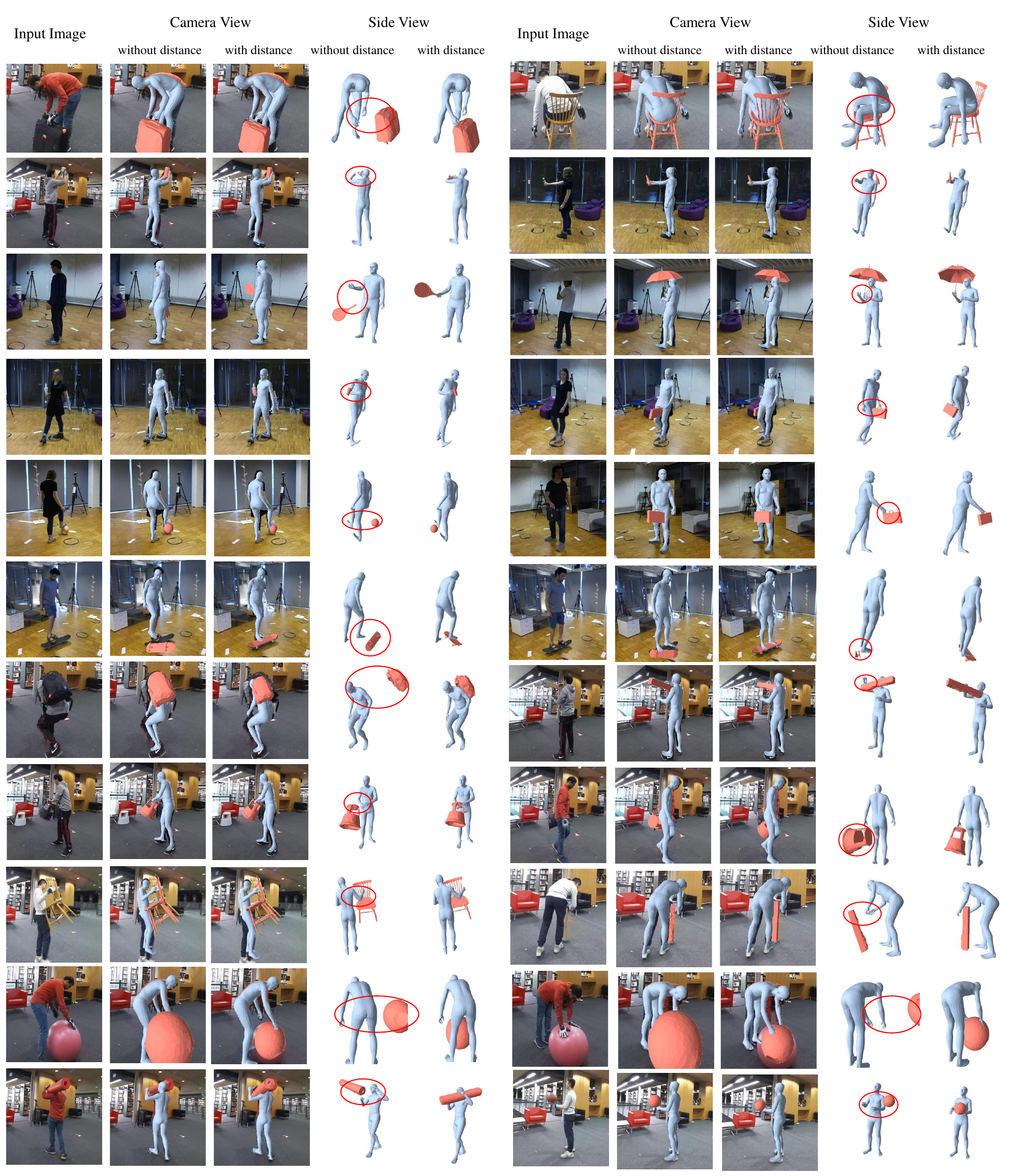}
		\caption{This figure shows the effectiveness of offset loss. If we align human and object with image in isolation, some inconsistent results will outcome. The offset loss can eliminate this happen in come cases.}
		\label{fig:comparison_distance}
	\end{figure*}

\begin{figure*}[htbp]
		\centering
		\includegraphics[width=\linewidth]{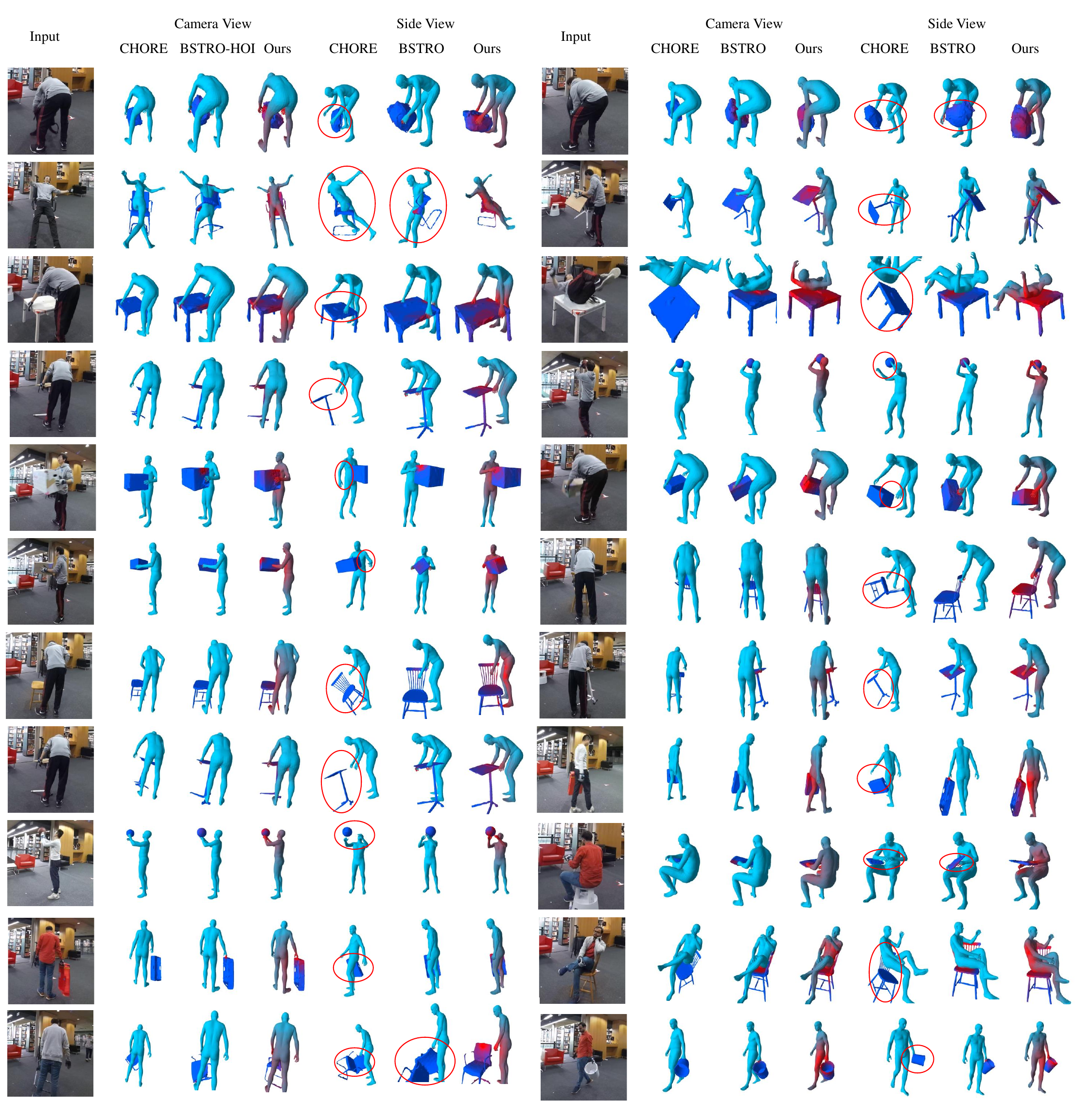}
		\caption{This figure shows comparisons between our method and other state-of-the-art methods qualitatively.}
		\label{fig:comparison_sota}
	\end{figure*}

    We show heavy occlusion cases in Figure \ref{fig:comparison_occlusion}. We compare our method with offset loss in post-optimization and without offset loss in post-optimization in Figure \ref{fig:comparison_distance}. We compare our method with other state-of-the-art method in Figure \ref{fig:comparison_sota}.
	
